\documentclass[11pt]{article}

\PassOptionsToPackage{sort, compress}{natbib}

\newif\ifarXiv
\arXivtrue

\usepackage{gtml2025_workshop}

\usepackage[english]    {babel}
\usepackage             {dsfont}
\usepackage             {fancyvrb}
\usepackage             {hyperref}
\usepackage             {graphicx}
\usepackage             {microtype}
\usepackage             {mathtools}
\usepackage             {subcaption}
\usepackage[svgnames]   {xcolor}
\usepackage[nameinlink] {cleveref}
\usepackage             {tcolorbox}

\ifarXiv
  \usepackage{listings}
\lstdefinestyle{bw}{
    basicstyle       = \ttfamily,
    keywordstyle     = \bfseries,
    commentstyle     = \itshape,
    stringstyle      = \ttfamily,
    showstringspaces = false,
    breaklines       = true,
  }
\else
  \usepackage{minted}
\fi

\VerbatimFootnotes

\definecolor{cardinal}{RGB}{196, 30, 58}
\definecolor{bleu}    {RGB}{ 49, 140, 231}

\tcbset{
  colframe  = bleu,
  colback   = bleu!10,
  fonttitle = \bfseries,
  boxsep    = 0.5mm,
  left      = 2mm,
  right     = 2mm,
sharp corners,
}

\hypersetup{
  colorlinks = true,
  urlcolor   = bleu,
  linkcolor  = bleu,
  citecolor  = bleu,
}

\title{Have Graph --- Will Lift?\\ The Case for Higher-Order Benchmarks}

\author{\textbf{Bastian Rieck}$^{1,2}$\\[6pt]
  $^{1}$AIDOS Lab, University of Fribourg, Switzerland\\
  $^{2}$Institute of AI for Health, Helmholtz Munich, Germany\\[4pt]
  \href{mailto:bastian.grossenbacher@unifr.ch}{\ttfamily bastian.grossenbacher@unifr.ch}
}

\date{}

\begin{document}

\maketitle

\begin{abstract}
  After a somewhat rocky start, geometry and topology have established
  a foothold in machine learning. Message passing, either on graphs or
  higher-order complexes, is one of the main drivers of \emph{geometric
  deep learning}, and paradigms that were once considered to be firmly
  in the realm of the abstract---like sheaves---have been ``tamed'' to
  serve as novel inductive biases for model architectures in
  \emph{topological deep learning}. The veritable diversity of models,
  however, is in stark contrast to the scarcity of suitable benchmark
  datasets. As a result, researchers often resort to \emph{lifting}
  existing graph datasets to include higher-order information. In this
  opinion paper, I want to encourage the community to also source new
  datasets, which may be used to prop up the foundations of our research
  field. 
\end{abstract}

\section{Introduction}

Graphs are an ubiquitous data modality, permitting researchers to model
complex phenomena in numerous domains~\citep{Velickovic23a}. For a long
time, the analysis of graphs, commonly referred to as \emph{graph
learning}, was dominated by \emph{graph kernels}, i.e., similarity
measures based on structural information such as the isomorphism class
of a graph~\citep{Morris23a, Borgwardt20}. In their seminal article,
\citet{Bronstein17a} outlined the field of \emph{geometric deep
learning}, which provides ``blueprints'' for developing models that are capable of
tackling data modalities like graphs and manifolds. The focus on
continuous and metric properties of data by geometric deep learning was
recently complemented by an algebraic and combinatorial perspective.
Methods of this sort are now commonly assigned to the field of \emph{topological
deep learning}~\citep{Bodnar22a}, which promises to improve relational
learning~\citep{Papamarkou24a}.

With both \emph{geometric deep learning}~(GDL) and \emph{topological deep
learning}~(TDL) making use of computational paradigms like \emph{message
passing}---albeit often on different modalities, i.e., graphs~(GDL) versus
higher-order data like simplicial complexes~(TDL)---the lines between
the fields become increasingly blurry. As such, attempts to provide
a precise separation are doomed to fail and this article will
refrain from making them. Instead, I want to focus on the way GDL/TDL
researchers \emph{analyze} their proposed methods. A typical evaluation
pipeline involves working with graph benchmark datasets and, depending
on whether one is interested in assessing the performance of
a higher-order method or not, \emph{lifting} such graphs to
a higher-order representation like a simplicial complex. Hence, instead
of working on the original graph~$\mathcal{G}$, one rather works on an associated
complex~$\mathcal{C}$. Any improvements in predictive performance are
then considered to reflect the positive influence of the proposed
method. While this approach is not wrong \emph{per se}, I want to
outline some of its shortcomings and, later on, propose alternatives, which
I believe would be generally beneficial for the GDL/TDL community at large.

\section{Quo Vadis, Graph?}

Lifting graphs to simplicial complexes or cellular complexes has been
a staple of TDL research since the beginning. In \citet{Bodnar21a},
graphs are lifted to simplicial complexes by building their \emph{clique
complex}, i.e., a simplicial complex in which each clique of~$k$
vertices is represented by a \mbox{$(k-1)$-dimensional} simplex. This is
later on extended to cellular complexes~\citep{Bodnar21b}. Other
complexes have been described in the literature, though, such as the 
\emph{neighborhood complex}~\citep[p.\ 130]{Matousek03a}, in which
simplices are subsets of vertices that share a common neighbor, or the
\emph{Dowker complex}~\citep{Dowker52a}, which requires the existence of
a binary relation defined on the vertices. Even more~(simplicial)
complexes are described by \citet{Jonsson08a}, typically with the goal
of providing insights into the topology of the underlying graphs and its
properties. In the context of \emph{topological data analysis}, lifting
is often conflated with the process of creating a \emph{filtration} of
a graph~(or complex), i.e., a nested way of indexing individual
substructures, leading to an independent rediscovery of some lifting
methods~\citep{Aktas19a}.

\begin{figure}[tbp]
  \centering
  \subcaptionbox{Clique complex}{\includegraphics[width=0.50\textwidth]{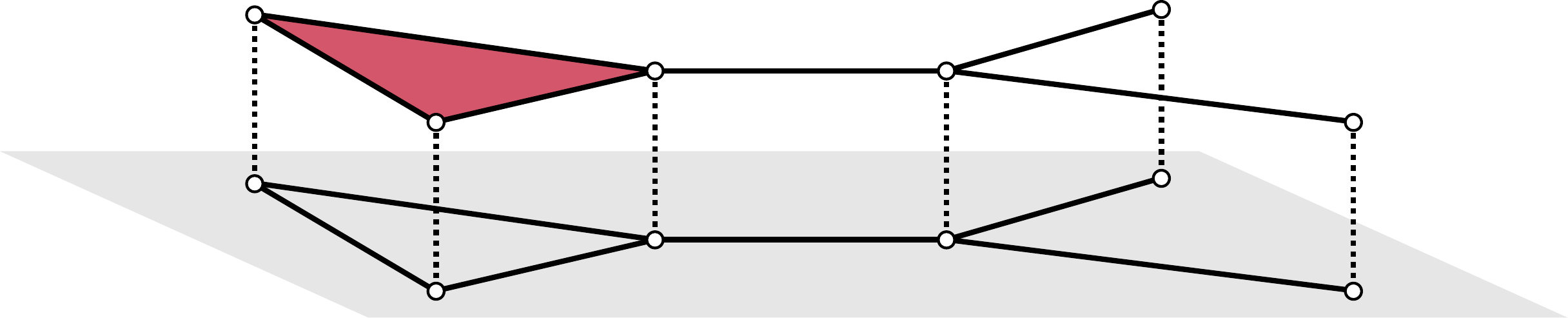}
  }\subcaptionbox{Neighborhood complex}{\includegraphics[width=0.50\textwidth]{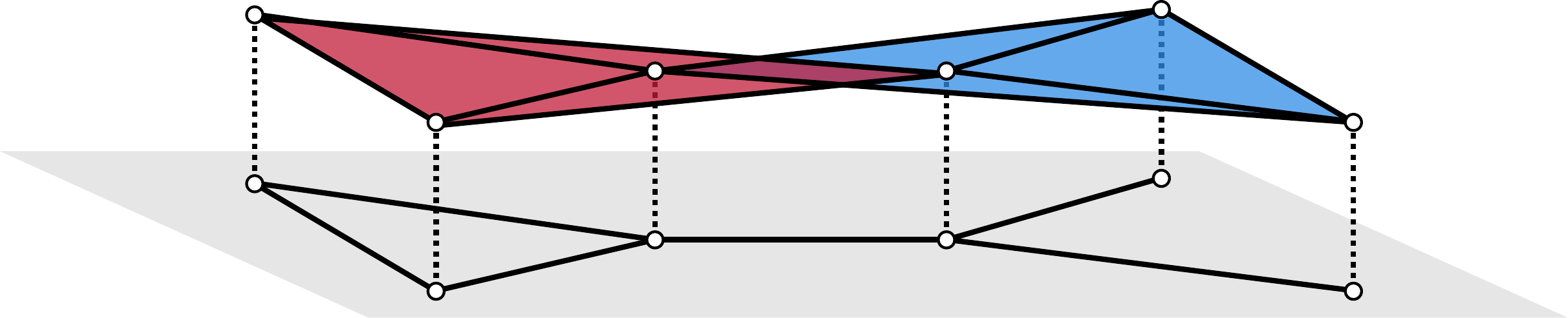}
  }\caption{An illustration of two different lifting strategies for a graph $G = (V, E)$.
    The \emph{clique complex}~$\mathcal{C}_{\mathrm{cl}}$ contains all
    subsets whose induced graphs are \emph{complete}, i.e.,
    $\mathcal{C}_{\mathrm{cl}} \coloneq \{ A \subseteq V \mid G[\sigma] \text{ is complete} \}$.
    The \emph{neighborhood complex}~$\mathcal{C}_{\mathrm{nb}}$, by contrast, contains all simplices
    whose vertices have a common neighbor, i.e.,
    $\mathcal{C}_{\mathrm{nb}} \coloneq \{ A \subseteq V \mid
    \mathrm{CN}(A) \neq \emptyset \}$, with $\mathrm{CN}(A) \coloneq
    \{v \in V \mid (v, a) \in E \text{ for all } a \in A\}$. The
    striking difference between the two constructions is that the
    clique complex contains only a single \mbox{$2$-simplex}, whereas
    the neighborhood complex contains two of them. In both cases, the
    original graph appears as the shadow of the respective complex; the
    ``right'' choice of lifting strategy is unclear a priori and depends
    on the dataset as well as the respective task. Lifting is thus an
    inherent part of the model selection process and should be treated
    as such.
  }
  \label{fig:Lifting example}
\end{figure}

\Cref{fig:Lifting example} depicts two lifting strategies. The resulting
complexes reflect different aspects of the underlying graph.
Hence, the lifting strategy needs to be treated as a crucial part of a
model and depends on the respective task. I cannot help but draw
a parallel to ``The Bitter Lesson''~\citep{Sutton19a} at this point: In
many cases, static lifting strategies seem to be hand-crafted
features, which, ultimately, may be replaced by a more general,
highly-scalable computational paradigm. This is not to say that
lifting strategies are not useful, though. In fact, the existence of
modern TDL libraries~\citep{Telyatnikov25a} can be a boon for improved
benchmarking, provided researchers are \emph{aware} of how to interpret
the choice of lifting strategy. It is unlikely that there is
a ``one-size-fits-all'' lifting procedure, hence any specific choice of
strategy should always be reported together with the model and may not
generalize. The best way to reason about lifting strategies is to
consider them as a part of a \emph{multiverse analysis}, i.e., an
analysis that explicitly encodes different choices~(such as the selection
of a lifting strategy or the selection of a model as well as the tuning of
hyperparameters), aiming to make research more transparent and more
reproducible~\citep{Steegen16a}. Our previous work found that latent
representations of generative models, for instance, are highly sensitive
with respect to their initial parameter choices~\citep{Wayland24a}. The
necessary ablation studies are, unfortunately, often absent from papers,
making it hard to assess the origin of gains in task performance.
In particular if the use of a lifting strategy results in a model with
more parameters, care must be taken so as not to confuse the source of
any observed improvements.

\begin{tcolorbox}[title = {In a nutshell}]
  For GDL/TDL applications using graph datasets, this
  implies that \emph{all} choices of lifting, including the
  \emph{absence} of a lifting strategy altogether, should be reported
  as ablations.
\end{tcolorbox}

This piece of advice is \emph{not} obviated by recent work, which
presents the first instance of \emph{learnable} lifting
strategies~\citep{Franco26a}. While learnable strategies are certainly
a step in the right direction, they need to be assessed critically like
any other crucial choice of model architecture. All of this
pontificating is not meant to downplay any of the numerous advances in
GDL/TDL. However, there have been some recent discussions concerning the
overall quality of our benchmark datasets and our research
culture~\citep{Bechler-Speicher25a}. More precisely, there is some
disconcerting evidence that the graph structure of many datasets is
largely \emph{irrelevant}, implying that task performance can be
improved by training on an empty graph or on a randomly-rewired
one~\citep{Coupette25a}. Put somewhat differently, we, as a community,
are just now starting to understand the intricate interplay between
graph \emph{structure} and graph \emph{features}. Moving to higher-order
domains will only exacerbate the complexity of this problem---we have
already seen evidence for this when extending \emph{message
passing} to simplicial complexes, for example~\citep{Taha25a}.

\begin{tcolorbox}[title = {In a nutshell}]
  Due to its influence on model design and training, lifting procedures
  may hide the \emph{relevance} of using higher-order information,
  especially for datasets in which the interplay between features and
  structure is not~(yet) well understood.
\end{tcolorbox}

\section{Ad Astra!}

Given that lifting may, in the worst case, hide the very thing we are
interested in studying, what are our options? I believe a viable path to
be the collection and construction of new datasets. This should make
gainful use of all the hard lessons gleaned from graph
learning. Inspired by the issues raised in \citet{Bechler-Speicher25a},
here are some suggestions:
\begin{enumerate}
  \item Datasets need to be constructed in a \emph{meaningful} way. The
    higher-order structure needs to be \emph{crucial} and
    \emph{advantageous} for the task at hand.

  \item Task \emph{complexity} should be known, for instance by
    including the performance of simple~(graph) baselines. This point is
    closely related to the previous one.

  \item Standard \emph{splits} should be available to ensure reproducibility
    and understand predictive performance. This should be accompanied by
    a certain minimum \emph{size} of the dataset to enable proper
    statistical reasoning.\footnote{Older graph datasets are often limited in size and the lack of
      predefined splits necessitates using techniques like
      cross-validation.
However, cross-validation needs to be \emph{repeated} to properly
      capture standard deviations~\citep{Errica20a, OBray21a}.
    }

  \item Datasets should be under \emph{version control} to enable their
    subsequent modification and error correction. While not always
    necessary, a proper \emph{maintenance plan} is required to account
    for community-led improvements.
\end{enumerate}

All of these items require a better understanding of the
\emph{provenance} of the data. That is, among other things, we need to
understand \emph{who} collected it for \emph{what} purpose and
\emph{how} it is supposed to be maintained. For many datasets in graph
learning, these questions cannot be reliably answered, unfortunately.
A quick analysis\footnote{Due to the way code in \texttt{pytorch-geometric} is structured, the
  following pattern will find most uses of dataset URLs:\begin{Verbatim}[frame = lines, vspace = 0pt]
rg -t py "^\s*url ="                               \
  | grep -Eo "(http|https)://[a-zA-Z0-9./?=_| awk -F/ '{print $3}'                           \
  | sort | uniq -c | sort -rn
\end{Verbatim}
} of patterns in the source code of
\texttt{pytorch-geometric}~\citep{Fey25a}, arguably one of the standard
software packages for graph learning, shows that datasets are hosted on
a variety of domains, including Amazon AWS~(5),
\texttt{dropbox.com}~(6), and the currently defunct
\texttt{graphmining.ai}~(6). Several datasets thus do not seem to be
directly under the control of the research community, setting us up
for a replication crisis in the future---even with the best intentions,
datasets might vanish because people move on to different jobs or delete
data by accident. There is no need for pessimism: Since this is an
open-source project, \emph{we}, as a research community, can make things
better. In the end, we will also benefit from it because we presumably
want to test our new models under the best conditions to make sure that
any signal we are receiving is actually part of the data and not just an
artifact.\footnote{The words of \citet{Feynman85a} should always be at the forefront of
  our mind:
\begin{quote}
    The first principle is that you must not fool yourself---and you are the easiest person to fool.
  \end{quote}}
If the recent upheavals in graph learning tells us only one thing, it
should be that we need to adopt a more critical approach towards
scholarship. \citet{Lipton18a} already mentioned general patterns that
we would do well to avoid, but in the age of large datasets, adopting
improved practices for ``data stewardship'' will only help us advance.
Better reporting like paper checklists or \emph{datasheets}~\citep{Gebru21a} can play a part here but only if we
treat the remainder of the data collection and curation process as
something worthwhile \emph{per se}.

\begin{tcolorbox}[title = {In a nutshell}]
  We need to treat \emph{data collection}, \emph{data curation}, and
  \emph{data maintenance} as integral steps of the model development
  process.
\end{tcolorbox}

\section{Acta Non Verba}

\begin{figure}[tbp]
  \centering
  \includegraphics[width=\linewidth]{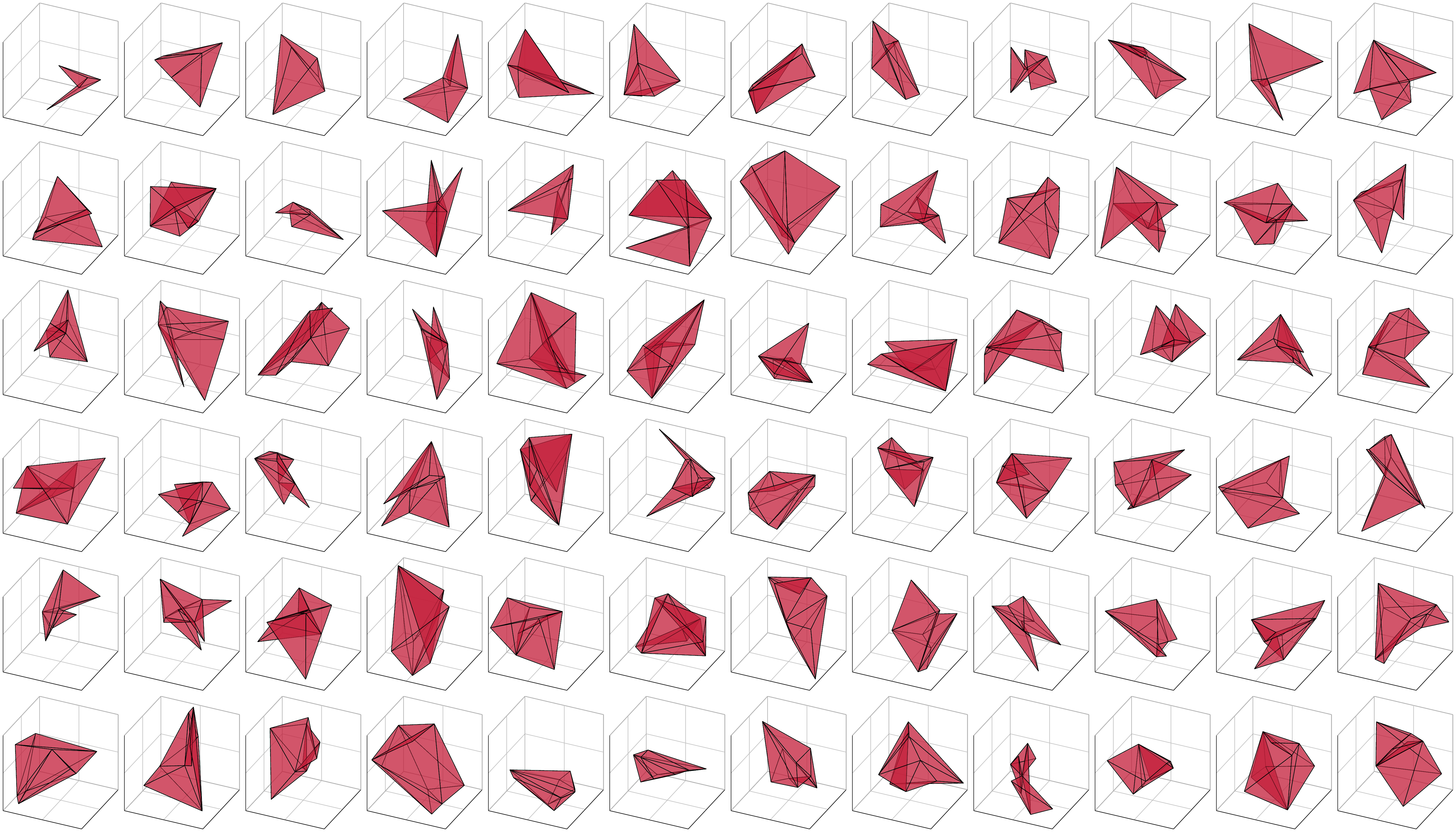}
  \caption{A visualization of some triangulations of $S^2$, the $2$-dimensional
    sphere, contained in the MANTRA dataset. The task of recognizing the
    topological type is well-defined but, due to the absence of
    \emph{canonical} coordinates, difficult for contemporary
    GDL/TDL models.
  }
  \label{fig:Triangulations}
\end{figure}

After pointing out issues, it is time for something more concrete and
savory. As a first step towards eventually establishing a large-scale
database of well-curated higher-order datasets, we started creating
MANTRA,\footnote{
  \url{https://github.com/aidos-lab/MANTRA}} the ``Manifold Triangulations Assemblage''~\citep{Ballester25a}.
MANTRA is based on the work of the late Frank H.\ Lutz~\citep{Lutz17a, Lutz08a},
who enumerated~(minimal) triangulations of \mbox{$2$-manifolds} and
\mbox{$3$-manifolds} and, where possible, provided a classification in
terms of topological invariants. This resulted in a vast dataset of more
than $43,000$ \mbox{$2$-manifolds} and more than $250,000$
\mbox{$3$-manifolds}, all of which are represented as \emph{abstract
simplicial complexes}. Unlike other datasets, these triangulations are
intrinsically higher-order---there is no lifting or prior modeling
involved. Moreover, the original data contains only combinatorial
information, i.e., there are no features for vertices, edges,
and simplices. Even in the case of \mbox{$2$-manifolds}, where
a \emph{random realization algorithm}~\citep{Lutz08a} can provide
coordinates, these triangulations have no ``canonical'' way of being
imbued with coordinates. As \Cref{fig:Triangulations} illustrates, the
random realizations do not help in recognizing the underlying manifold,
thus increasing dataset complexity.
On the bright side, however, an abstract, coordinate-free representation
lends itself to a simple textual description. Here is an example entry
of the database we curated, showing one particular triangulation of the
\mbox{$2$-sphere} as well as some associated properties:

\ifarXiv
\begin{lstlisting}[language = {}]
  {
    "id": "manifold_2_4_5_1",
    "triangulation": [[1,2,3], [1,2,4], [1,3,4], [2,3,4]],
    "dimension": 2,
    "n_vertices": 4,
    "betti_numbers": [1, 0, 1],
    "torsion_coefficients": ["", "", ""],
    "name": "S^2",
    "genus": 0,
    "orientable": true,
    "vertex_transitive": true
  }
\end{lstlisting}

\clearpage

\else
\begin{minted}{json}
  {
    "id": "manifold_2_4_5_1",
    "triangulation": [[1,2,3], [1,2,4], [1,3,4], [2,3,4]],
    "dimension": 2,
    "n_vertices": 4,
    "betti_numbers": [1, 0, 1],
    "torsion_coefficients": ["", "", ""],
    "name": "S^2",
    "genus": 0,
    "orientable": true,
    "vertex_transitive": true
  }
\end{minted}
\fi

Some of this information is \emph{redundant}---for instance,
\emph{orientability} can also be determined purely based on the last
Betti number---or \emph{specific} to \mbox{$2$-manifolds}~(like the
\emph{genus}). However, even in this abstract form, the dataset gives
rise to \emph{regression} or \emph{classification} tasks:
\begin{enumerate}
  \item Predict Betti numbers or torsion coefficients~(regression).
  \item Predict orientability, vertex transitivity, or canonical name~(classification).
\end{enumerate}
These tasks can be addressed in a myriad of ways, making, for instance, use of
higher-order message passing methods~\citep{Telyatnikov25a},
exploiting graph representations like the \mbox{$1$-skeleton}, or
employing transformer-like structural encodings~\citep{Ballester24a, Carrasco25a}.

\paragraph{Are these tasks hard?}
However, in light of the issues raised above, we should first think about what we
are trying to accomplish here---these tasks appear somewhat
``natural,'' but what about their utility and complexity? Here, we need
to go on a brief tangent: Topologists have fully classified~(closed)
\mbox{$2$-manifolds} already. Every closed \mbox{$2$-manifold} is known
to be homeomorphic to either $S^2$, a \emph{connected sum} of $k$ tori,
i.e., $\#^k T^2$, or a \emph{connected sum} of $k$ projective planes,
i.e., $\#^k \mathds{RP}^2$. Consequently, the \emph{Euler characteristic} and
\emph{orientability} are sufficient to classify
a \mbox{$2$-manifold}. Thus, for \mbox{$2$-manifolds}, the tasks
described above serve as an important ``smoke test'' of overall model
quality: Any \emph{topological deep learning} model should aim to
exhibit perfect predictive performance here. 
Success would indicate that the model is
sufficiently \emph{expressive} to learn basic topological invariants.
The situation for \mbox{$3$-manifolds} is more complex---while the
classification based on canonical name is simple for the current version
of the dataset, there is still value in models trying to predict other
properties~(Betti numbers, torsion coefficients), \emph{despite} the
fact that homology algorithms solve them.
Moreover, there is no need to stop at dimension~$3$. Even though
higher-dimensional manifolds are known to be impossible to
classify~\citep{Kreck14a}, a large-scale dataset could help address
research-level questions for topologists. While we are not there yet,
I am convinced that if our TDL models remain incapable of going well
beyond simplicial homology calculations, we are likely experiencing
\emph{fundamental challenges} in model design that we have yet to
overcome.

\paragraph{The performance of contemporary models.}
Our preliminary experiments~\citep{Ballester25a} indicate that both GDL
and TDL models struggle to solve the aforementioned tasks. While TDL
models~(specifically, those making use of higher-order message passing)
exhibit overall slightly better performance than their GDL
counterparts~(which are by necessity restricted to a graph
representation of the data, i.e., the \mbox{$1$-skeleton}), even the
case of \mbox{$2$-manifolds} is far from being solved. And this is not
even scratching the surface of the transformations that we have at our
disposal when dealing with triangulations. For instance, classical
constructions like \emph{barycentric subdivision} do not change the
homeomorphism type of a manifold. Nevertheless, when subdividing
a triangulation, we observe performance decreases. The~(sobering)
conclusion of these experiments is that the current generation of TDL
models are \emph{not} making use of the inherent topology of the data but
rather perform local aggregations on combinatorial structures. As such,
they are focusing on a \emph{combinatorial} prior that is not
aligned with the \emph{topological} equivalence relation required to
solve the tasks.

\paragraph{We are still at the beginning.}
I strongly believe that this should only be the \emph{beginning} of the
story, however. To better understand the failure modes of our models, we need
other datasets of the same ilk, i.e., with known provenance,
well-defined tasks, and intrinsically higher-order, both \emph{with} and
\emph{without} node coordinates. It is my hope that, through community
contributions, MANTRA may become a repository of many diverse
triangulations, a higher-order companion to collections like
GraphBench~\citep{Stoll26a}. MANTRA already provides versioning
information, release tracking, and many additional conveniences that ensure
proper maintenance and reproducibility---everyone is cordially invited
to participate!

\section{Conclusion}

Having hopefully demonstrated the necessity and utility of new and
extended benchmarks, I want to raise some open questions, which may
serve as inspiration for advancing the field:
\begin{enumerate}
  \item How can we move \emph{beyond} expressivity measures borrowed from
    graph learning? We recently looked into how structural measures that
    explicitly incorporate feature information could be
    defined~\citep{Carrasco26a}; an equivalent for higher-order data
    would seem appropriate and useful.
  \item What other~(manifold-based) datasets would be appropriate as
    benchmarks? Can we find some that are \emph{intrinsically} equipped
    with coordinates/features?
  \item Which \emph{other} invariants or properties do we want to predict? Even
    in the case of \mbox{$3$-manifolds}, computational topology has
    developed new algorithms to measure numerous properties that are
    characteristic of a triangulation or its underlying manifold~\citep{Burton04a, Burton25a}.
\end{enumerate}
Whatever we decide to do, we should ensure that TDL methods are not
merely \emph{inspired} by topological concepts but are also equipped
with strong topological guarantees.

\subsubsection*{Acknowledgments}
I am grateful for the opportunity to discuss these issues with the
community and extend heartfelt thanks to the organizers of the Workshop
on Geometry, Topology, and Machine Learning~(GTML 2025) for
accommodating me. The opinions in this article have been honed in many
discussions with my collaborators, in particular Mathieu Alain, Rubén
Ballester, Nello Blaser, Ernst Röell, Martin Carrasco, Corinna Coupette, Kelly Maggs,
Johannes S.\ Schmidt, Emily Simons, and Jeremy Wayland. Any errors in judgment are solely my
responsibility.
This work has received funding from the Swiss State Secretariat for
Education, Research, and Innovation~(SERI).

\clearpage

\bibliographystyle{gtml2025_workshop}
\bibliography{main}

\end{document}